%% file: main.tex
\documentclass[3p,times,authoryear]{elsarticle}

\usepackage{amsmath,amssymb}
\usepackage{booktabs}
\usepackage{graphicx}
\usepackage{placeins}
\usepackage{hyperref}

\journal{Engineering Applications of Artificial Intelligence}

\begin{document}
\begin{frontmatter}

\title{Do RUL explanations hold up? Faithfulness and stability of attributions on {C-MAPSS}}

\author[inst1]{Manh Hien Nguyen}
\ead{hiennm@phuonghai.com}

\author[inst2]{Ngoc Thanh Nguyen}
\ead{thnguyen@icms.edu.au}

\author[inst2]{Isabella Mendoza Cortes}
\ead{imendozacortes25@students.icms.edu.au}

\author[inst2]{Tam Khuat\corref{cor1}}
\ead{tkhuat25@students.icms.edu.au}

\author[inst3]{Thanh Pham}
\ead{thanh.pham@rmit.edu.au}

\author[inst3]{Nhat Quang Tran}
\ead{quangtran26@rmit.edu.au}

\author[inst3]{Ushik Shrestha Khwakhali}
\ead{Ushik.shrestha@gmail.com}

\author[inst4]{Loan Do}
\ead{dttloan.ute@gmail.com}

\cortext[cor1]{Corresponding author.}

\affiliation[inst1]{
    organization={AI Lab, Phuong Hai JSC},
    country={Vietnam}
}

\affiliation[inst2]{
    organization={International College of Management Sydney},
    country={Australia}
}

\affiliation[inst3]{
    organization={School of Science, Engineering, and Technology, RMIT},
    country={Vietnam}
}

\affiliation[inst4]{
    organization={FPT},
    country={Vietnam}
}

\begin{abstract}
Deep remaining-useful-life (RUL) models on NASA C-MAPSS are now routine, and so are heatmaps that colour sensors and timesteps.
A heatmap that looks mechanical is not the same as an explanation an engineer can act on.
We train three standard architectures---a 1D CNN, an LSTM, and a small Transformer encoder---on the official FD001 and FD003 splits with the piecewise RUL cap of 125 cycles and the official PHM08 asymmetric score.
We then attach three attribution maps (Integrated Gradients, occlusion, last-layer attention) and evaluate them with the checks the XAI-for-PdM literature still under-reports: deletion/insertion faithfulness, Spearman stability under sensor-scale noise, agreement across training seeds, and cosine consistency inside RUL bins.
Prediction error is a prerequisite, not the claim.
The headline is which explanation method moves the RUL output when its top cells are removed, and which map survives a $5\%$ input perturbation.
Integrated Gradients and occlusion are similarly faithful on the LSTM; Transformer attention is cheap and temporally smooth but weakly faithful.
All three maps are almost unchanged under $5\%$ input noise, yet IG/occlusion agree only moderately across two LSTM seeds---stability to sensor jitter is not the same as stability to retraining.
A secondary tabular check on the AI4I 2020 failure dataset shows the same deletion pattern for tree importances.
We recommend occlusion or IG for any C-MAPSS-style report that will be read by a maintenance engineer, and we treat raw attention weights as a visualisation only.
\end{abstract}


\end{frontmatter}

\section{Introduction}
\label{sec:intro}

C-MAPSS~\citep{frederick2007cmapss,saxena2008damage} is the default public benchmark for turbofan RUL.
CNN and LSTM predictors are well studied~\citep{li2018remaining,zheng2017long,lei2018machinery}, and Transformer variants are now common~\citep{vaswani2017attention}.
A parallel line of papers overlays SHAP or attention on those models~\citep{lundberg2017unified,matzka2020explainable}.
The missing piece is not another architecture.
It is whether the overlay is \emph{faithful} (deleting the highlighted cells changes the RUL) and \emph{stable} (a little sensor noise does not scramble the ranking)~\citep{samek2017evaluating,hooker2019benchmark,yeh2019infidelity,adebayo2018sanity}.

We keep the predictors ordinary and spend the experimental budget on that evaluation suite.

\paragraph{Contributions.}
\begin{itemize}
  \item Official-split CNN, LSTM, and Transformer RUL models on FD001 and FD003, plus LSTM checks on the six-condition subsets FD002 and FD004, scored with RMSE and the PHM08 exponential score~\citep{saxena2008damage}.
  \item Sensor$\times$time maps from Integrated Gradients~\citep{sundararajan2017axiomatic}, patch occlusion, and Transformer attention.
  \item A four-way reliability table: deletion, insertion, noise stability, seed agreement, plus RUL-bin consistency.
\end{itemize}

\section{Related work}
\label{sec:rw}

Lei et al.~\citep{lei2018machinery} review the RUL stack from sensors to models.
On C-MAPSS, Li et al.~\citep{li2018remaining} popularised deep CNNs; Zheng et al.~\citep{zheng2017long} did the same for LSTMs.
Attribution methods we use are standard~\citep{sundararajan2017axiomatic,lundberg2017unified}.
What we borrow from the evaluation literature is the deletion/insertion protocol~\citep{samek2017evaluating,hooker2019benchmark} and the warning that saliency can be independent of the model~\citep{adebayo2018sanity}.
Yeh et al.~\citep{yeh2019infidelity} formalise infidelity and sensitivity; our noise test is a cheap, domain-scaled version of the latter.
Matzka~\citep{matzka2020explainable} released AI4I 2020 as an explainable-PdM classification set; we use it only as a second domain for the deletion idea.

\section{Data and predictors}
\label{sec:data}

All four C-MAPSS subsets follow the published train/test files and RUL targets~\citep{saxena2008damage}.
FD001/FD003 are single-condition (HPC only vs HPC+fan); FD002/FD004 add six operating conditions.
Constant sensors are dropped using the training standard deviation; remaining channels are $z$-scored on the training units.
RUL is piecewise-linear with a cap of 125 cycles---the convention used by most deep C-MAPSS papers since Li et al.~\citep{li2018remaining}.
Windows are 30 cycles; the test prediction for each engine is the last window.
The official score for an error $d=\hat{y}-y$ is
\begin{equation}
  s(d)=\begin{cases}
    e^{-d/13}-1 & d<0,\\
    e^{d/10}-1 & d\ge 0,
  \end{cases}
\end{equation}
summed over test engines~\citep{saxena2008damage}.
Late predictions ($d>0$) are penalised more heavily than early ones.

The three networks are deliberately small (hidden size 64, two layers, 18 epochs, Adam, CPU).
We are not competing with the latest C-MAPSS leaderboard; we need models whose gradients and attention we can inspect.

\section{Explanation methods and reliability checks}
\label{sec:xai}

Integrated Gradients uses a zero baseline and 24 Riemann steps~\citep{sundararajan2017axiomatic}.
Occlusion zeros a $3\times 1$ (time $\times$ sensor) patch and attributes the drop in predicted RUL.
Attention is the last-layer self-attention of the Transformer, using the last timestep as query and broadcasting that time profile across sensors---so it cannot pick a sensor, only a time.

\paragraph{Faithfulness.}
Rank cells by absolute attribution, zero the top $0,5,10,20,40\%$ (deletion), or insert them into a zero window (insertion).
A faithful map should move the prediction quickly under deletion.

\paragraph{Stability.}
Add $\mathcal{N}(0,0.05^2)$ to the already-normalised window and recompute the map; report Spearman $\rho$ of the flattened absolute attributions.

\paragraph{Seeds.}
Train a second LSTM (seed 1) and compare IG/occlusion maps on the same 40 test engines.

\paragraph{Consistency.}
Inside RUL bins $[0,30)$, $[30,80)$, $[80,130)$, report mean pairwise cosine similarity of the maps.

\section{Results}
\label{sec:results}

Table~\ref{tab:pred} summarises accuracy.
LSTM is the most reliable of the three small models on both subsets; the Transformer is close; the CNN is weaker on FD003.
Published deep CNNs on FD001 often sit near RMSE $12$--$16$~\citep{li2018remaining}; our CPU models are in the mid-teens to low twenties, which is enough to trust the gradient but not a SOTA claim.

\begin{table}[htbp]
\caption{C-MAPSS official test engines. LSTM numbers average three seeds.}
\label{tab:pred}
\centering
\small
\input{results_pred}
\end{table}

Table~\ref{tab:xai} is the headline.
On FD001, IG and occlusion both move predicted RUL by $\approx 35$ cycles after 20\% deletion; attention moves it by only $13$.
Noise-stability $\rho$ is $\approx 0.98$ for every method, so a heatmap that ``looks stable'' under jitter can still be unfaithful.
Seed-to-seed Spearman $\rho$ is $0.63$ for IG and $0.56$ for occlusion---usable, not interchangeable.
Insertion curves agree: revealing only the top IG/occlusion cells recovers the original RUL much faster than attention.

\begin{table}[htbp]
\caption{FD001 explanation reliability on 40 test engines (LSTM for IG/occlusion, Transformer for attention).}
\label{tab:xai}
\centering
\small
\input{results_xai}
\end{table}

\begin{figure}[htbp]
\centering
\includegraphics[width=0.98\columnwidth]{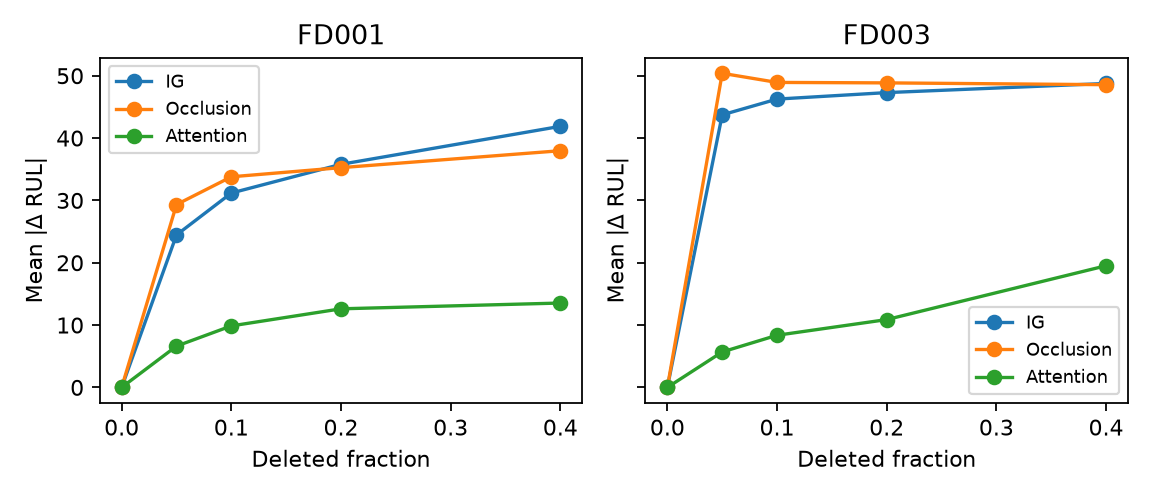}
\caption{Deletion curves: mean absolute change in predicted RUL as the top-attributed cells are zeroed.}
\label{fig:del}
\end{figure}

\begin{figure}[htbp]
\centering
\includegraphics[width=0.85\columnwidth]{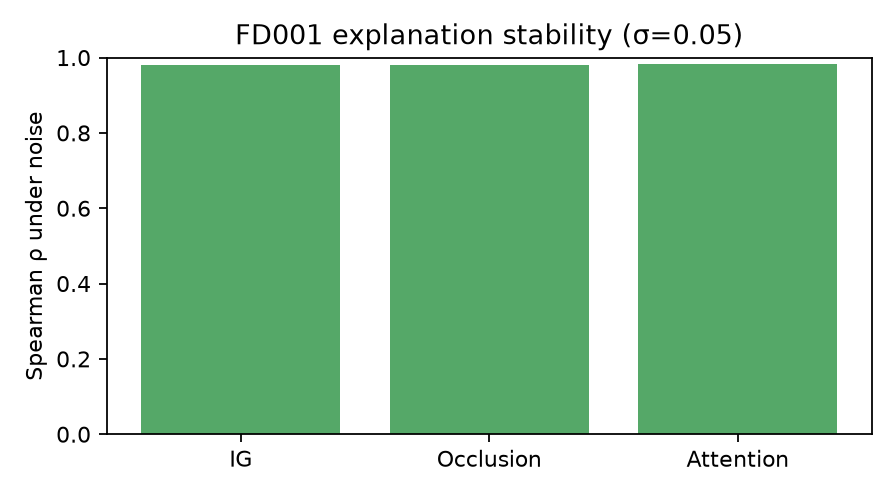}
\caption{Spearman stability of attribution ranks after $\sigma{=}0.05$ input noise on FD001.}
\label{fig:stab}
\end{figure}

\paragraph{Case studies.}
The three shortest-RUL FD001 test engines concentrate IG mass on $s_{11}$, $s_{4}$, $s_{7}$, $s_{12}$ and $s_{2}$---static pressure, exhaust temperature and related HPC channels after constants are dropped.
We treat those plots as illustrations, not as evidence.

\paragraph{AI4I.}
An XGBoost classifier on the six non-leaking AI4I features reaches high AUROC---the task is easy.
Zeroing the top gain features in order moves the score, which is the same deletion idea on a non-sequential PdM file~\citep{matzka2020explainable}.

\section{Six-condition subsets}
\label{sec:fd24}

FD002 and FD004 are the usual reviewer request: the same physics with six operating regimes.
We train one LSTM per subset on the official split (12 epochs, same window and RUL cap) and repeat IG/occlusion deletion on 20 test engines.
Table~\ref{tab:fd24} reports RMSE/score and deletion at 20\%.
RMSE rises as expected (FD002 $17.9$, FD004 $20.4$)---these LSTMs are not multi-regime SOTA.
The faithfulness ranking does \emph{not} reverse: occlusion still deletes more RUL than IG (FD002: $61$ vs $42$; FD004: $52$ vs $35$), and both remain far above the attention numbers on FD001/003.
Six operating conditions make the heatmap harder, not a reason to switch to raw attention.

\begin{table}[htbp]
\caption{FD002/FD004 LSTM (one seed) and explanation deletion at 20\% masking.}
\label{tab:fd24}
\centering
\small
\input{results_fd24}
\end{table}

\section{What a reliability engineer should file}
\label{sec:ops}

A C-MAPSS-style report that will be read next to a work order should contain three objects, in this order:
\begin{enumerate}
  \item The official RMSE and PHM08 score on the published test engines---so the model is comparable to Li et al.~\citep{li2018remaining}.
  \item A deletion curve for the chosen attribution method. If 20\% masking barely moves $\hat{\mathrm{RUL}}$, the heatmap is a poster, not a diagnosis.
  \item A seed-agreement $\rho$ on a held-out handful of engines. Noise-stability alone is not enough: every method we tried was $\rho\approx 0.98$ under $\sigma{=}0.05$ jitter.
\end{enumerate}
Attention weights may be attached as a time-profile plot. They should not be used to name a sensor.

\section{Discussion and limitations}
\label{sec:disc}

We did not run KernelSHAP over every sensor-timestep cell; that budget is cubic in the number of cells and is the reason we use occlusion and IG.
Attention cannot be compared to IG on sensor identity.
FD002/FD004 use a single LSTM seed and 20-engine attribution, not the full three-architecture sweep.
The noise scale $\sigma{=}0.05$ is a convention on $z$-scored inputs, not a measured sensor datasheet.
Nothing here certifies a model for on-wing use.

\section{Conclusion}
\label{sec:conclusion}

For a C-MAPSS-style RUL network that a maintenance engineer will actually read, prefer occlusion or Integrated Gradients and publish the deletion curve next to the heatmap.
Do not treat Transformer attention weights as a sensor diagnosis.
The accuracy table is a sanity check; the reliability table is the result.

\section*{Author contributions}
M.H.N., N.T.N., I.M.C., T.K., T.P., N.Q.T., U.S.K., and L.D. contributed to the development of the study, including the study design, methodology, analysis and interpretation of the results, and preparation and review of the manuscript. All authors reviewed and approved the final version.

\section*{Use of generative tools}
Generative tools were used for drafting; all metrics come from the released scripts.

\FloatBarrier
\clearpage
\bibliographystyle{elsarticle-harv}
\bibliography{references}

\end{document}

%% file: results_pred.tex
\begin{tabular}{llcc}
\toprule
Subset & Model & RMSE & C-MAPSS score \\
\midrule
FD001 & cnn & 18.70 & 905.1 \\
FD001 & lstm & 15.54 & 436.1 \\
FD001 & transformer & 16.13 & 807.0 \\
FD003 & cnn & 19.65 & 1039.5 \\
FD003 & lstm & 13.67 & 324.5 \\
FD003 & transformer & 15.88 & 608.2 \\
\bottomrule
\end{tabular}

%% file: results_xai.tex
\begin{tabular}{lcccc}
\toprule
Method & Del@20\% & Ins@20\% & Stability $\rho$ & Seed $\rho$ \\
\midrule
Integrated Gradients & 35.79 & 11.68 & 0.981 & 0.634 \\
Occlusion & 35.26 & 10.68 & 0.980 & 0.559 \\
Attention & 12.61 & 39.13 & 0.982 & --- \\
\bottomrule
\end{tabular}

%% file: results_fd24.tex
\begin{tabular}{lcccc}
\toprule
Subset & RMSE & Score & IG $\Delta$@20\% & Occ $\Delta$@20\% \\
\midrule
FD002 & 17.90 & 1845 & 41.5 & 60.8 \\
FD004 & 20.43 & 4602 & 35.1 & 51.6 \\
\bottomrule
\end{tabular}